\documentclass[a4paper]{article}
\usepackage{amssymb}
\usepackage{amsmath}
\usepackage{amsthm}
\usepackage[authoryear,sectionbib]{natbib}
\usepackage{hyperref} 
\usepackage[pdftex]{graphicx,color}
\usepackage{authblk}

\theoremstyle{definition}
\newtheorem{remark}{Remark}

\title{Solving the Goddard problem\\ by an influence diagram\thanks{This work 
was supported by the Czech Science Foundation (project 16-12010S).}}
\author{Ji\v{r}\'{\i} Vomlel and V\'{a}clav Kratochv\'{\i}l}
\affil{Institute of Information Theory and Automation,\\
Czech Academy of Sciences, \\
Pod vod\'arenskou v\v{e}\v{z}\'{\i} 4, Prague 8, 182 08, Czechia\\
vomlel@utia.cas.cz,\\
\texttt{http://www.utia.cas.cz/vomlel/}}
\date{}

\begin{document}
\maketitle

\begin{abstract}
Influence diagrams are a decision-theoretic extension of probabilistic graphical models.
In this paper we show how they can be used to solve the Goddard problem. 
We present results of numerical experiments with this problem and
compare the solutions provided by influence diagrams with the optimal solution.
\end{abstract}

\section{Introduction}

Formulated by Robert H. Goddard, see~\citep{goddard-1919}, the problem is to
establish the optimal thrust profile for a rocket ascending vertically 
from the Earth's surface to achieve a given altitude with a given speed 
and pay load and with the minimum fuel expenditure.
The aerodynamic drag and the gravitation varying with the altitude are considered.
We study a problem version that assumes a bounded thrust.
The problem has become a benchmark in the optimal control theory due to
a characteristic singular arc behavior in connection with a
relatively simple model structure, which makes the Goddard
problem an ideal object of study, c.f.~\citep{graichen-petit-2008}.

\section{The ODE model}

The equation of motion of the rocket, subject to the forces of gravity, drag,
and thrust are:
\begin{eqnarray}
 c \cdot \dfrac{d m(t)}{d t} + m \cdot  \dfrac{d v(t)}{d t} + d(v,h)
+ m \cdot g(h) & = & 0  \label{eq-1} \enspace ,
\end{eqnarray}
where
\begin{eqnarray}
d(v,h) & = & \frac{1}{2} \cdot s \cdot c_D \cdot  \rho(h) \cdot  v^2 \label{eq-2} \\
\rho(h) & = & \rho_0 \cdot \exp\left(\beta \cdot \left(1-\dfrac{h}{R}\right)\right) \label{eq-3} \\
g(h) & = & g_0 \cdot \dfrac{R^2}{h^2} \label{eq-3b}
\end{eqnarray}
using the following notation:
\begin{itemize}
\item $c$ is the exhaust velocity (the jet speed),
\item $m$ is the rocket mass composed from the pay load $m_p$, which is a constant, 
and the fuel $m_f$, which is burnt during the rocket ascent, 
\item $m_0$ is the initial rocket mass,
\item $t$ is the time,
\item $v$ is the speed of the rocket, 
\item $h$ is the altitude measured as the distance from the Earth's center, 
\item $R=6371$ km is the radius of the Earth,
\item $s$ is the cross-section area of the rocket, 
\item $d(v,h)$ is the drag at speed $v$ and altitude $h$,
\item $c_D$ is the dimensionless drag constant, 
\item $\rho(h)$ is the density of the air at altitude $h$, 
\item $\rho_0$ is the density of the air at the Earth's surface,
\item $\beta$ is a dimensionless constant, 
\item $g(h)$ is the acceleration of gravity at altitude $h$, and
\item $g_0=9.81 m/s^2$ is the gravitational acceleration at the Earth's surface.
\end{itemize}
If we describe the system dynamics with respect to the altitude
the formula~\eqref{eq-1} transforms to:
\begin{eqnarray}
c \cdot v \cdot \dfrac{d m}{d h} + m \cdot v \cdot \dfrac{d v}{d h} + d(v,h) + m \cdot g(h) & = & 0 
\label{eq-4} \enspace .
\end{eqnarray}

\section{Normalized Goddard Problem}

In literature the problem is often presented in its nondimensional form.
Let $G=g_0 \cdot R^2$. 
We use tilde to denote the new nondimensionalized variables:
\begin{eqnarray*}
\tilde{m} & = & \dfrac{m}{m_0} \\
\tilde{h} & = & h \cdot \dfrac{1}{R} \\
\tilde{t} & = & t \cdot \sqrt{\dfrac{G}{R^3}} \enspace .
\end{eqnarray*}
This leads to 
\begin{eqnarray*}
\tilde{v} & = & v \cdot \sqrt{\dfrac{R}{G}} \\
\tilde{a} & = & a \cdot \dfrac{R^2}{G} \enspace .
\end{eqnarray*} 
The nondimenzionalized model transformed from formulas~\eqref{eq-4} and~\eqref{eq-2}--\eqref{eq-3b} 
is the following:
\begin{eqnarray}
\tilde{c} \cdot  \sqrt{\dfrac{G}{R}} \cdot m_0 \cdot \sqrt{\dfrac{G}{R^3}} \cdot \dfrac{d \tilde{m}}{d \tilde{h}} \cdot \tilde{v}
+ \tilde{m} \cdot m_0 \cdot \sqrt{\dfrac{G}{R}}  \cdot \sqrt{\dfrac{G}{R^3}} \cdot  \dfrac{d \tilde{v}}{d \tilde{h}} \cdot \tilde{v}\nonumber \\
+ d(\tilde{v},\tilde{h})
+ \tilde{m} \cdot m_0 \cdot g(\tilde{h}) & = & 0  \label{eq-1-nd} \enspace ,
\end{eqnarray}
where
\begin{eqnarray}
d(\tilde{v},\tilde{h}) & = & \frac{1}{2} \cdot \tilde{s} \cdot R^2 \cdot c_D \cdot  \rho(\tilde{h}) \cdot  
\left(\tilde{v} \cdot  \sqrt{\dfrac{G}{R}}\right)^2 \label{eq-2-nd} \\
\rho(\tilde{h}) & = & \tilde{\rho}_0 \cdot \dfrac{m_0}{R^3} \cdot \exp\left(\beta \cdot \left(1-\tilde{h}\right)\right) \label{eq-3-nd} \\
g(\tilde{h}) & = & \dfrac{G}{R^2} \cdot \dfrac{R^2}{\tilde{h}^2 \cdot R^2} 
 \ \ = \ \ \dfrac{G}{R^2} \cdot\dfrac{1}{\tilde{h}^2} \label{eq-3b-nd} \enspace .
\end{eqnarray}
By substituting~\eqref{eq-3-nd} to~\eqref{eq-2-nd} we get:
\begin{eqnarray}
d(\tilde{v},\tilde{h}) & = & \frac{1}{2} \cdot \tilde{s} \cdot c_D \cdot  \tilde{\rho}_0 \cdot \dfrac{m_0 \cdot G}{R^2} \cdot \exp\left(\beta \cdot \left(1-\tilde{h}\right)\right) \cdot  
\tilde{v}^2 \label{eq-2-nd2} 
\end{eqnarray}
and by substituting~\eqref{eq-2-nd2} and~\eqref{eq-3b-nd} to~\eqref{eq-1-nd} and dividing both sides of the equation 
by $\dfrac{m_0 \cdot G}{R^2}$ we get:
\begin{eqnarray}
\tilde{c} \cdot   \dfrac{d \tilde{m}}{d \tilde{h}} \cdot \tilde{v}
+ \tilde{m} \cdot  \dfrac{d \tilde{v}}{d \tilde{h}} \cdot \tilde{v}
+ \frac{1}{2} \cdot \tilde{s} \cdot c_D \cdot  \tilde{\rho}_0 
\cdot \exp\left(\beta \cdot \left(1-\tilde{h}\right)\right) \cdot  \tilde{v}^2 
+  \dfrac{\tilde{m}}{\tilde{h}^2} & = & 0  \label{eq-1-nd2} \enspace , \rule{6mm}{0mm}
\end{eqnarray}

\begin{remark}
In the sequel we will use the normalized Goddard Problem. 
For simplicity, we will omit tildes.
\end{remark}


\section{Optimal control problem formulation}

The state variables are the rocket mass $m$ (of pay load and fuel) 
and the rocket speed $v$ at altitude $h$. 
The control variable $u$ controls the engine thrust, which is the derivative of mass $m$ with respect to time $t$
multiplied by the jet speed $c$, i.e.
\begin{eqnarray}
 u & = & c \cdot \dfrac{d m}{d t}  \ \ = \ \  c \cdot \dfrac{d m}{d h} \cdot \dfrac{d h}{d t} \ \ = \ \  \dfrac{d m}{d h} \cdot c \cdot v \enspace .\label{eq-mass-0}
\end{eqnarray}
which implies that the mass $m$ at the altitude $h$ is
\begin{eqnarray}
m(h) & = &  m_0 + \int_{h'=0}^{h} c \cdot v(h') \cdot u(h') \ \ dh'
\enspace ,\label{eq-mass}
\end{eqnarray}
where $m_0$ is the initial mass at the rocket launch.
Please note that $u \leq 0$ due to the fact that the mass of the rocket can only decrease
(by burning the fuel). The control will be restricted to $u \in [-3.5,0]$.

The task is to find a control function $u(h)$ so that we get from the initial state $(m_0,v_0)$ 
to a terminal state $(m_T,v_T)$, where $m_T$ is the terminal mass 
and $v_T$ is the terminal speed at a given terminal altitude $h_T$, 
$v_0$ is the initial speed, and $m_0 > m_T$ is the initial rocket mass (including fuel)
so that the with a minimal fuel consumption (i.e, with a maximal final mass).

Formula~\eqref{eq-1-nd2} can be rewritten as:
\begin{eqnarray}
u \ + \ m \cdot v \cdot \dfrac{d v}{d h} 
\ + \ m \cdot \dfrac{1}{h^2} 
\ + \ \frac{1}{2} \cdot s \cdot c_D \cdot  \rho_0 \cdot \exp\left(\beta \cdot \left(1-h\right)\right) \cdot  v^2 
& = & 0 \enspace . \rule{8mm}{0mm} \label{eq-5} 
\end{eqnarray}

The formula~\eqref{eq-mass-0} can be written using 
a newly defined function $g(h,v)$ and formula~\eqref{eq-5} 
using a newly defined function $f(h,v)$ as:
\begin{eqnarray}
\dfrac{dm}{dh} & = & g(u,v) \ \ = \ \ \dfrac{u}{c \cdot v} \label{eq-mass-2}\\
\dfrac{d v}{d h} & = & f(h,m,u,v) \label{eq-6} \\
& = & - \frac{u}{m \cdot v} \ - \ \frac{1}{2 \cdot m} \cdot s \cdot c_D \cdot  \rho_0 \cdot \exp\left(\beta \cdot \left(1-h\right)\right) \cdot  v
\ - \ \dfrac{1}{v \cdot h^2} \nonumber \enspace .
\end{eqnarray}
This equation we will use for the derivation of approximate methods in the next section.

The control will be restricted to $u \in [-3.5,0]$.
It is assumed that the rocket is initially at rest at the surface of the Earth
and that its fuel mass is 40\% of the rocket total mass.
The initial and terminal values will be
\begin{eqnarray*}
h_0 & = & 1 \\
h_T & = & 1.01 \\
m_0 & = & 1 \\
v_0 & = & 0 \\
m_T & \geq & 0.6 \cdot m_0 \ \ = \ \ 0.6 \enspace .
\end{eqnarray*}

\subsection{Model parameters}

The model parameters we consider correspond to those presented in~\citep{tsiotras-kelley-1991} and~\citep{seywald-cliff-1992}. 
The aerodynamic data and the vehicle's parameters originate from~\citep{zlatskiy-kiforenko-1983} and 
correspond roughly to the Soviet SA-2 surface-to-air missile, NATO code-named Guideline.
The nondimensionalized values of these constants are:
\begin{eqnarray*}
\beta & = & 500 \\
s \cdot \rho_0 & = & 12400 \\
c_D & = & 0.05 \\
c & = & 0.5 \enspace .
\end{eqnarray*}

\section{ODE approximate solution methods\label{sec-ode}}

Now, consider a trajectory segment of length $\Delta h$ with the control being constant and equal to $u'$. 
Let $h'$ be the altitude, $m'$ the mass value, and $v'$ the speed -- all at the beginning of the segment.

\subsection*{The Euler method}

In the Euler method the following approximation is used:
\begin{eqnarray}
\left(\begin{array}{l}
m(h'+\Delta h) \\
v(h'+\Delta h) 
\end{array}\right) 
& \approx & 
\left(\begin{array}{l}
m' + \Delta h \cdot g(u',v') \\[3mm]
v' + \Delta h \cdot f(h',m',u',v') 
\end{array}\right) 
\label{eq-euler}\enspace .
\end{eqnarray}
This method is not very accurate. Note that in the mass estimation
a constant rocket speed is assumed for the whole segment, which means that during speed-ups
the mass is underestimated and during slow-down the mass is overestimated.
Also, note that in the speed estimation a constant rocket mass and a constant drag is assumed 
for the whole segment, which means the speed is underestimated.
On the other hand the method is extremely fast.

\subsection*{The general approximation method}

The general approximation of order $s$ takes the following form:
\begin{eqnarray}
\left(\begin{array}{l}
m(h'+\Delta h)\\
v(h'+\Delta h)
\end{array}\right) 
& \approx & 
\left(\begin{array}{l}
m(h') + \Delta h \cdot \sum_{i=1}^s w_i \cdot \ell_i \\[2mm]
v(h') + \Delta h \cdot \sum_{i=1}^s w_i \cdot k_i 
\end{array}\right) \label{eq-runge-kutta} \enspace , \rule{5mm}{0mm} \label{eq-approx}
\end{eqnarray}
where for $i=1,\ldots,s$ 
\begin{eqnarray}
\ell_i & = & g\left(
\begin{array}{l}
u', \\ 
v' + \Delta h \cdot \sum_{j=1}^s a_{i,j} \cdot k_j 
\end{array}
\right) \label{eq-general-1} \\[2mm]
k_i & = & f\left(
\begin{array}{l}
h'+z_i \cdot \Delta h, \\
m'+ \Delta h \cdot \sum_{j=1}^s a_{i,j} \cdot \ell_j, \\
u', \\
v'+\Delta h \cdot \sum_{j=1}^s a_{i,j}\cdot k_j 
\end{array}
\right) \enspace , \rule{5mm}{0mm} \label{eq-general-2}
\end{eqnarray}
which reduces to one equation for each $i=1,\ldots,s$:
\begin{eqnarray}
k_i 
& = &  f\left(
\begin{array}{l}
h'+z_i \cdot \Delta h, \\
m'+ \Delta h \cdot \sum_{j=1}^s  \dfrac{a_{i,j} \cdot u'}{c} \cdot \left(v' + \Delta h \cdot \sum_{\ell=1}^s a_{j,\ell} \cdot k_{\ell} \right)^{-1},	\\
u', \\ 
v'+\Delta h \cdot \sum_{j=1}^s a_{i,j}\cdot k_j 
\end{array}
\right)  \enspace . \rule{5mm}{0mm} \label{eq-general-2b}
\end{eqnarray}

\subsection*{The classical Runge--Kutta method}

In the the classical Runge--Kutta method of order $s=4$ (RK4) 
the coefficients' values are given by the following Butcher tableau:
\begin{center}
\begin{tabular}{c|cccc}
$z_1$ & $a_{1,1}$ & $a_{1,2}$ & $a_{1,3}$ & $a_{1,4}$ \\[2mm]
$z_2$ & $a_{2,1}$ & $a_{2,2}$ & $a_{2,3}$ & $a_{2,4}$ \\[2mm]
$z_3$ & $a_{3,1}$ & $a_{3,2}$ & $a_{3,3}$ & $a_{3,4}$ \\[2mm]
$z_4$ & $a_{4,1}$ & $a_{4,2}$ & $a_{4,3}$ & $a_{4,4}$ \\[2mm]
\hline\\[-3mm]
& $w_1$ & $w_2$ & $w_3$ & $w_4$
\end{tabular}
\ \ = \ \ 
\begin{tabular}{c|cccc}
$0$ & $0$ & $0$ & $0$ & $0$ \\[2mm]
$\frac{1}{2}$ & $\frac{1}{2}$ & $0$ & $0$ & $0$ \\[2mm]
$\frac{1}{2}$ & $0$ & $\frac{1}{2}$ & $0$ & $0$ \\[2mm]
$1$ & $0$ & $0$ & $1$ & $0$ \\[2mm]
\hline\\[-3mm]
& $\frac{1}{6}$ & $\frac{2}{6}$ & $\frac{2}{6}$ & $\frac{1}{6}$
\end{tabular}
\end{center}

The computational advantage of RK4 is that (due to the zeroes in its Butcher tableau) 
the values of $\ell_i$ and $k_i$ for 
$i=1,\ldots,s$ are specified explicitly.
Unfortunately, for some problems the RK4 method can be numerically unstable 
unless the step size is extremely small.
This may lead to wild oscillations of the control. 

\subsection*{Gauss--Legendre method}

The Butcher tableau of this method for $s=2$ is
\begin{center}
\begin{tabular}{c|cc}
$z_1$ & $a_{1,1}$ & $a_{1,2}$ \\[2mm]
$z_2$ & $a_{2,1}$ & $a_{2,2}$ \\[2mm]
\hline\\[-3mm]
& $w_1$ & $w_2$ 
\end{tabular}
\ \ = \ \ 
\begin{tabular}{c|cc}
$\frac{1}{2}-\frac{1}{6}\sqrt{3}$ & $\frac{1}{4}$ & $\frac{1}{4} - \frac{1}{6}\sqrt{3}$ \\[3mm]
$\frac{1}{2}+\frac{1}{6}\sqrt{3}$ & $\frac{1}{4}+\frac{1}{6}\sqrt{3}$ & $\frac{1}{4}$ \\[3mm]
\hline\\[-3mm]
& $\frac{1}{2}$ & $\frac{1}{2}$ 
\end{tabular}
\end{center}
Note that the values of $k_i$ for $i=1,\ldots,s$ are specified only implicitly and
the non-linear system specified by~\eqref{eq-general-2b} must be solved.
When their values are found they can be substituted to formula~\eqref{eq-general-1}~and~\eqref{eq-approx}.
Contrary to RK4 the Gauss--Legendre method is A-stable~\citep{dahlquist-1963}.

\subsection*{Control constraints}

The rocket jets cannot produce an infinite force, 
which implies that the absolute value\footnote{Note that $u \leq 0$.} 
of control $u$ is restricted from above. The upper bound is assumed 
to be constant during the whole flight:
\begin{eqnarray}
|u| & \leq &  |u_{max}| \enspace . \label{eq-upper-bound} 
\end{eqnarray}

To avoid situations the rocket is not moving 
or falling down we require its speed $v > 0$.
Also, the rocket cannot have its mass lower than is its payload, i.e., $m \geq m_p$.
We realize these constraints by means of control restrictions. We allow only control
values $u$ for which, when they are substituted to formula~\eqref{eq-runge-kutta},
it holds that  
\begin{eqnarray*}
m(h'+\Delta h) & \geq & m_p \ \ \mbox{and}\\ 
v(h'+\Delta h) & > & 0 
\end{eqnarray*}
except the terminal altitude $h_T$ where $v(h_T) \geq 0$.
Note that the lower bound of $|u|$ is thus a function of the altitude $h$, the current 
speed $v$, and the current rocket mass $m$.

\section{The influence diagram}

\begin{figure}[htbp]
\begin{center}
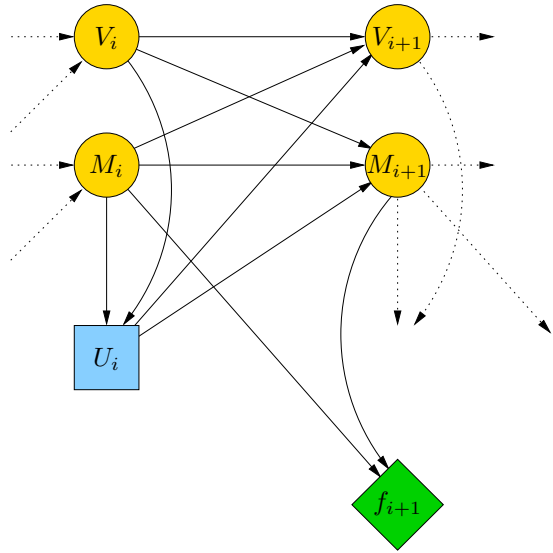
\caption{A Segment of the Influence Diagram for the Goddard Problem\label{fig-id-goddard}}
\end{center}
\end{figure}

In each segment $i$ ($i = 0,1,\ldots,N$) of of the influence diagram for the Goddard Problem, there are two state variables:
\begin{itemize}
\item a speed variable $V_i$ and 
\item a mass variable $M_i$. 
\end{itemize}
In each segment $i$ there is also one decision variable:
\begin{itemize} 
\item the control of the thrust of the rocket engine $U_i$.
\end{itemize}
Finally, in each segment $i = 1,\ldots,N$ one utility node is present:
\begin{itemize} 
\item the fuel consumption in the segment $f_i$.
\end{itemize}
The structure of one segment of the influence diagram for the discrete version 
of the Goddard Problem is presented in Figure~\ref{fig-id-goddard}.

In each segment a solution of a system of two ordinary differential equations
is found by an approximate method as it is discussed in Section~\ref{sec-ode}. 
Typically, the computed mass and speed values at the end of the segment 
will not lay in the discrete set of values of the mass and speed variables.
Therefore we will approximate the state transformations by non-deterministic CPTs 
$P(V_{i+1}|V_i,M_i)$ and $P(M_{i+1}|V_i,M_i)$ 
as it is described in~\citep[Section 5.2]{kratochvil-vomlel-2016}.

\section{Experimental results}

\begin{figure}[htbp]
\begin{center}
\includegraphics[width=0.8\textwidth]{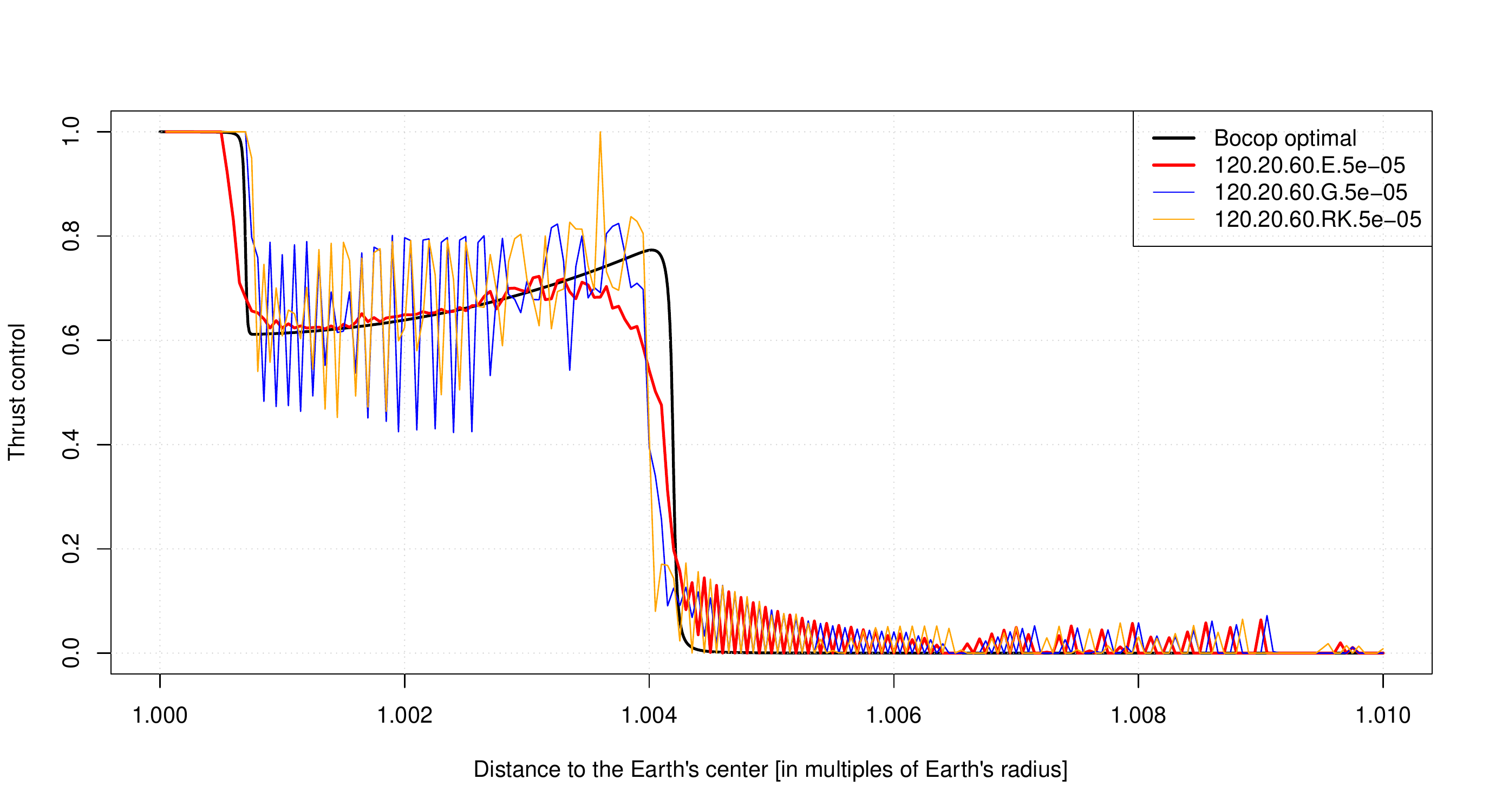}
\includegraphics[width=0.8\textwidth]{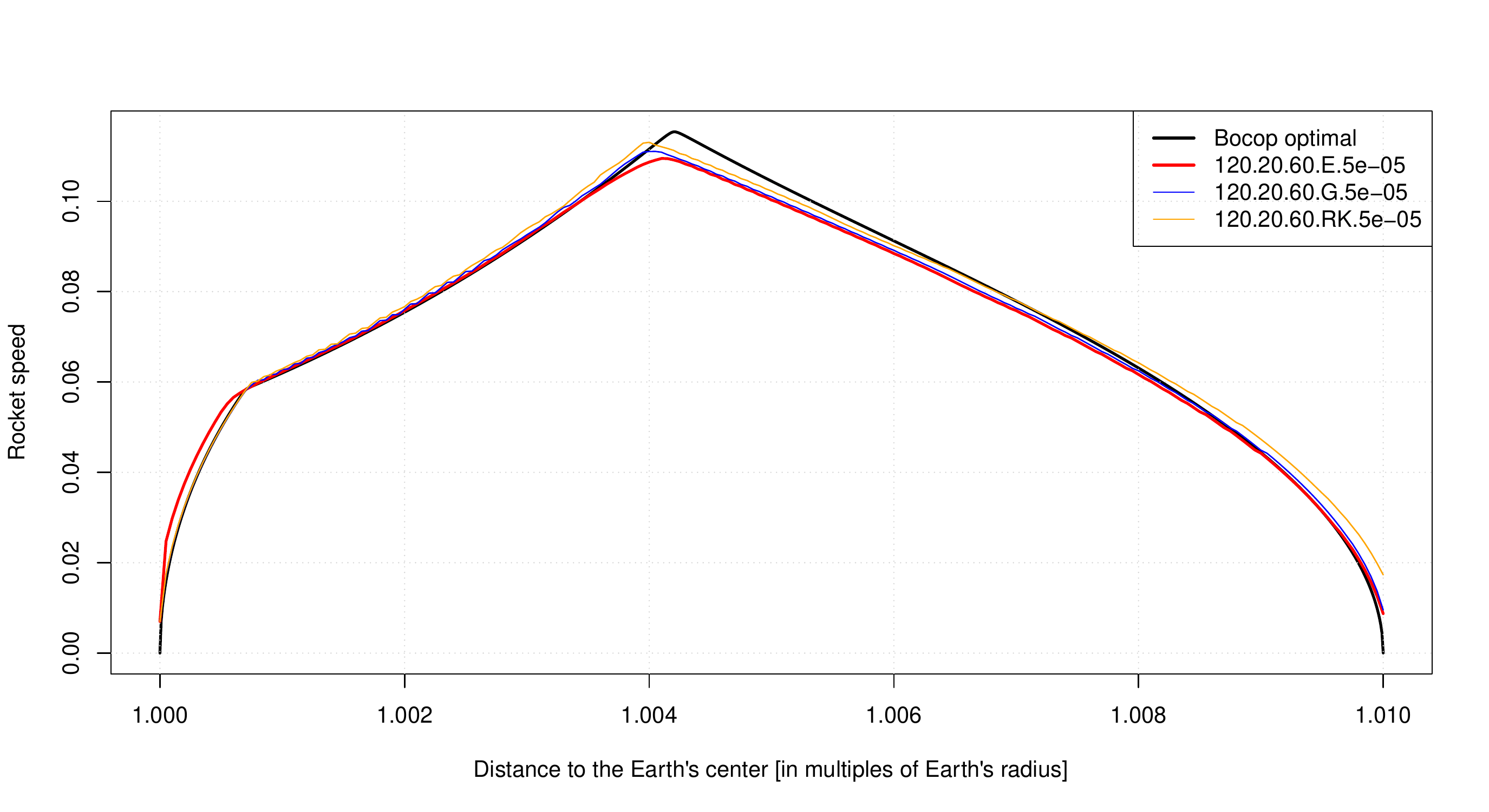}
\includegraphics[width=0.8\textwidth]{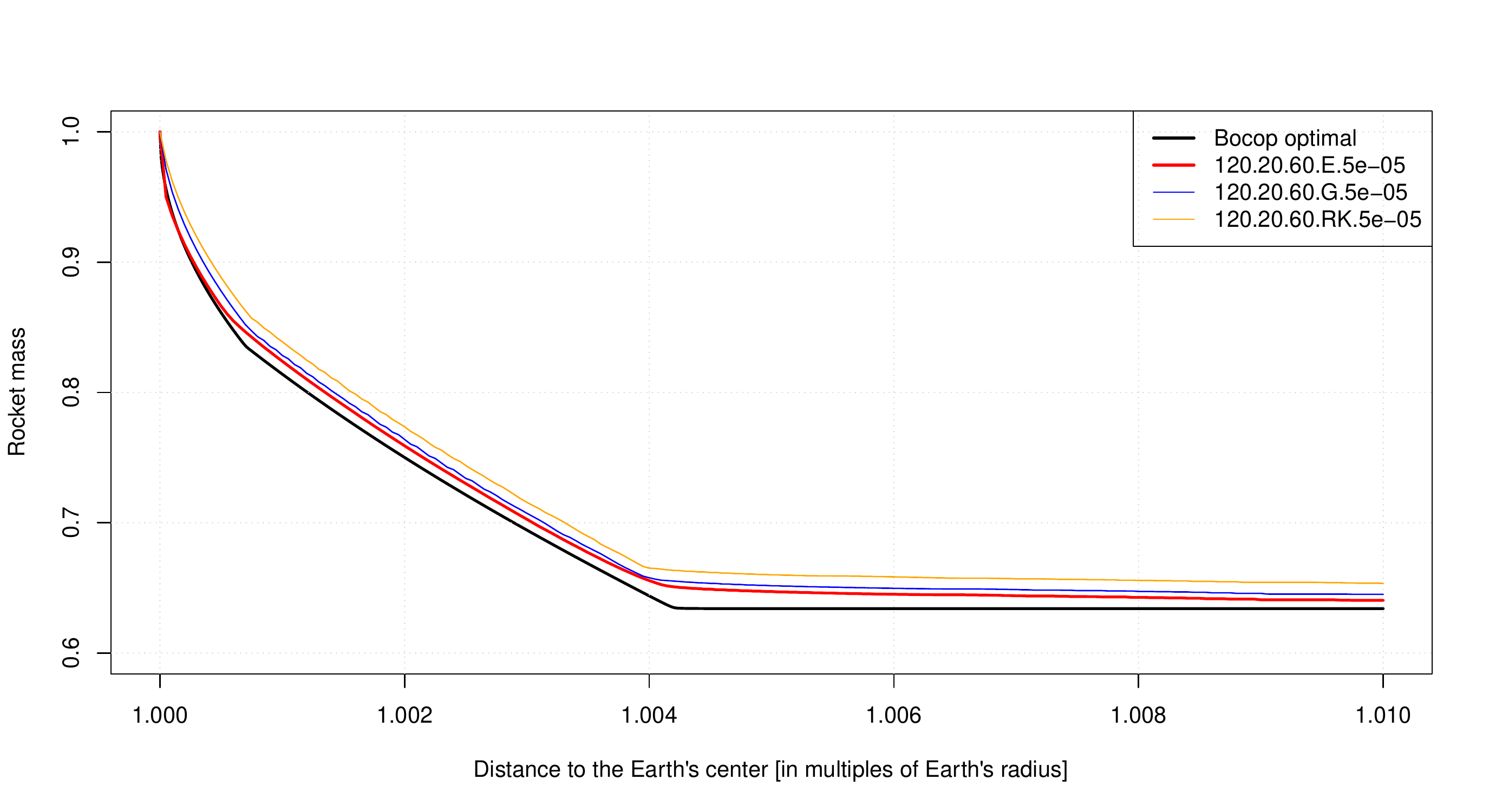}
\end{center}
\caption{Comparisons of the optimal solution with influence diagram solutions.\label{fig-compare-goddard}}
\end{figure}

In Figure~\ref{fig-compare-goddard} we compare the control, speed, and mass profiles of the 
optimal solution found by Bocop~\citep{bocop-2016} with solutions found by influence diagrams with different
discretizations and different approximation methods.
It is known~\citep{miele-1963} that the optimal solution consists of three
subarcs: (a) a maximum-thrust subarc, (b) a variable-thurst subarc, and 
(c) a coasting subarc, i.e., a subarc with the zero thrust.

We denoted the solutions found by influence diagrams using a name schema 
{\tt v.u.m.M.h} composed from the parameters used in the experiments:
\begin{itemize}
\item {\tt v} ... the number of states of the speed variables,
\item {\tt u} ... the number of states of the control variables,
\item {\tt m} ... the number of states of the mass variables,
\item {\tt M} ... the discretization method for solving ODEs ({\tt E} stands for the Euler method,
{\tt RK} for the Runge-Kutta method, and 
{\tt G} for the Gauss--Legendre method), and
\item {\tt h} ... the length of the trajectory segment.
\end{itemize}

By looking at Figure~\ref{fig-compare-goddard} we can conclude that 
the Euler method best approximates optimal control and suffers from smaller oscillations of the control. 
The control strategies found by the Runge-Kutta and the Gauss-Legendre methods have larger oscillations.
The speed and the mass profiles are similar for all methods 
and they are close to the optimal profiles found by BOCOP.

\section{Conclusions}

We have shown how influence diagrams can be used to solve a control theory benchmark problem 
-- the Goddard Problem. The numerical experiments reveal that the solution found by influence diagrams 
approximates well the optimal solution and quality of approximation improves with finer discretizations.
From the tested ODE approximation methods the best results were achieved by the simplest one -- the Euler
method.

\bibliographystyle{apalike}
\bibliography{bibliography}

\end{document}